\documentclass[letterpaper, 10 pt, conference]{ieeeconf}  % Comment this line out if you need a4paper

\IEEEoverridecommandlockouts                              % This command is only needed if 
\usepackage{bm}
\usepackage{url}
\usepackage{amsmath}
\usepackage{nccmath}
\usepackage{comment}
\usepackage{balance}
\usepackage{multirow}
\usepackage{textcomp}
\usepackage{graphicx}
\usepackage{lipsum}
\usepackage{hyperref}
\usepackage{algorithm}
\usepackage{dblfloatfix}
\usepackage[noend]{algpseudocode}
\usepackage[margin=1pt, font=footnotesize, skip=2pt]{subcaption}

\title{\LARGE \bf
Adaptive Hyperparameter Tuning for Black-box LiDAR Odometry
}

\author{Kenji Koide$^{1}$, Masashi Yokozuka$^{1}$, Shuji Oishi$^{1}$, and Atsuhiko Banno$^{1}$% <-this % stops a space
\thanks{$^{1}$ All the authors are with the Department of Information Technology and Human Factors, the National Institute of Advanced Industrial Science and Technology, Umezono 1-1-1, Tsukuba, 3050061, Ibaraki, Japan, {\tt\small k.koide@aist.go.jp}}%
}

\newcommand{\argmax}{\mathop{\rm arg~max}\limits}
\newcommand{\argmin}{\mathop{\rm arg~min}\limits}

\begin{document}

\maketitle
\thispagestyle{empty}
\pagestyle{empty}

%%%%%%%%%%%%%%%%%%%%%%%%%%%%%%%%%%%%%%%%%%%%%%%%%%%%%%%%%%%%%%%%%%%%%%%%%%%%%%%%
\begin{abstract}
This study proposes an adaptive data-driven hyperparameter tuning framework for black-box 3D LiDAR odometry algorithms. The proposed framework comprises offline parameter-error function modeling and online adaptive parameter selection. In the offline step, we run the odometry estimation algorithm for tuning with different parameters and environments and evaluate the accuracy of the estimated trajectories to build a surrogate function that predicts the trajectory estimation error for the given parameters and environments. Subsequently, we select the parameter set that is expected to result in good accuracy in the given environment based on trajectory error prediction with the surrogate function. The proposed framework does not require detailed information on the inner working of the algorithm to be tuned, and improves its accuracy by adaptively optimizing the parameter set. We first demonstrate the role of the proposed framework in improving the accuracy of odometry estimation across different environments with a simulation-based toy example. Further, an evaluation on the public dataset KITTI shows that the proposed framework can improve the accuracy of several odometry estimation algorithms in practical situations.
\end{abstract}

%%%%%%%%%%%%%%%%%%%%%%%%%%%%%%%%%%%%%%%%%%%%%%%%%%%%%%%%%%%%%%%%%%%%%%%%%%%%%%%%
\section{INTRODUCTION}

%%%%%%%%%%%%%%%%%%%%%%%%%%%%%%%%%%%%%%%%%%%%%%%%%%%%%%%%%%%%%%%%%%%%%%%%%%%%%%%%

\begin{figure*}[tb]
  \centering
  \includegraphics[width=0.8\linewidth]{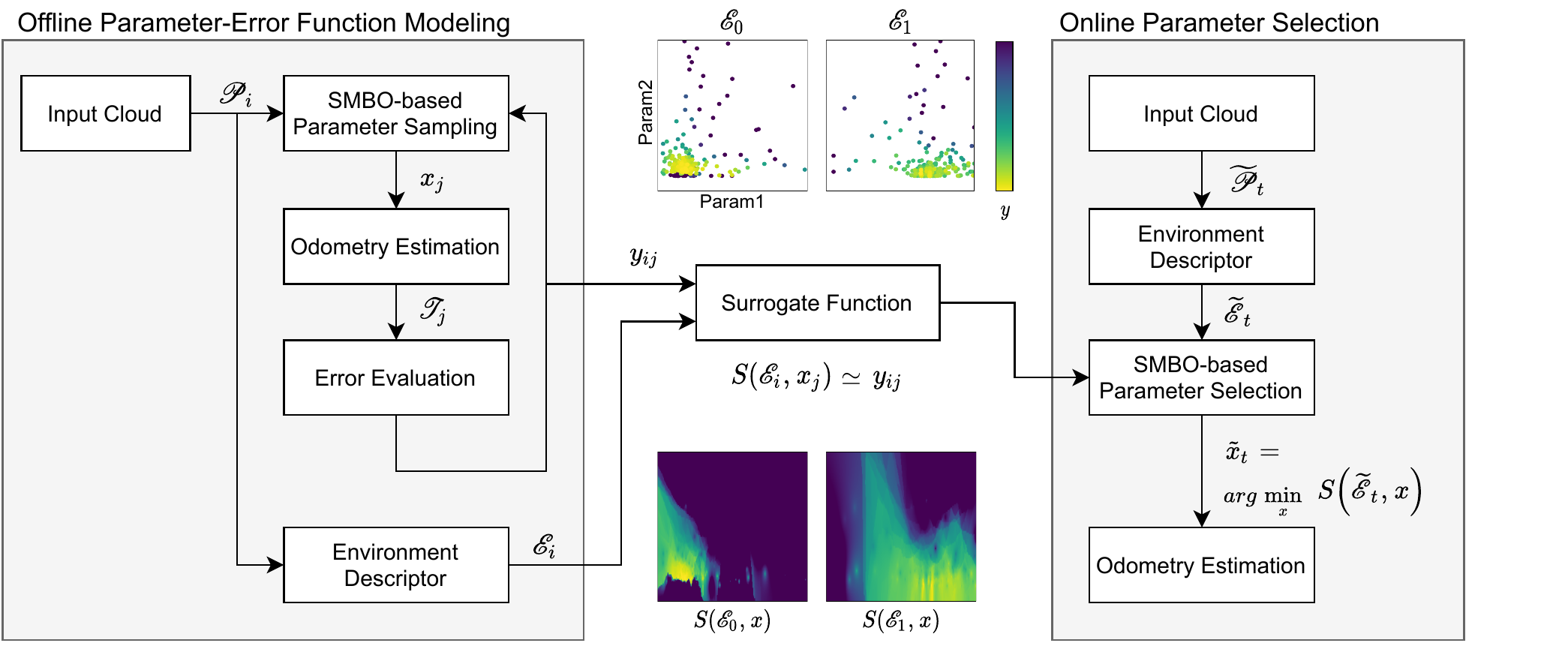}
  \caption{Overview of the proposed adaptive parameter tuning framework.}
  \label{fig:system}
\end{figure*}

LiDAR odometry estimation is inevitable for autonomous systems such as mobile service robots and autonomous driving vehicles. A number of LiDAR odometry estimation and mapping algorithms have been proposed, and many autonomous systems are built based on these robust and accurate algorithms. However, these algorithms are often surprisingly complex and involve many hyperparameters. Their performance can be significantly affected by the choice of hyperparameters; therefore, the hyperparameters must be fine-tuned depending on the system and environment. Despite their significance, hyperparameters are manually tuned in most studies, which results in considerable human effort and suboptimal parameter sets. Although a few studies have proposed automatic parameter tuning methods for odometry estimation \cite{Pomerleau_2011, Nobili_2017}, they are all dedicated to specific algorithms, sensors, and task configurations \cite{Pomerleau2015}.

In our previous work, we proposed a general and automatic hyperparameter optimization approach for {\it black-box} odometry estimation methods \cite{koide_icra2021}. This approach repeats the parameter-sampling-and-trial loop with efficient parameter space exploration based on sequential model-based optimization (SMBO). It uses only information on the input parameter set and output trajectory of the algorithm to be tuned (i.e., {\it black-box} modeling) and facilitates the improvement of the performance of any odometry estimation methods without detailed information on their inner working. Although this approach is effective in improving the performance of odometry estimation methods on certain metrics and datasets, we observed that it suffers considerably from overfitting to the training sequences. If the training set contains only easy sequences, the SMBO tends to choose a very aggressive parameter set that results in instability in other sequences. Conversely, if there is a difficult sequence in the training set, the optimized parameter set tends to be very conservative, resulting in deteriorated trajectory accuracy.

In this work, to deal with overfitting and achieve a balance between robustness and accuracy, we propose a data-driven approach for adaptively selecting the hyperparameters of {\it black-box} odometry estimation methods. The proposed framework automatically models the effects of the differences in hyperparameters on the behavior of the odometry estimation algorithm, and then selects the best parameter set for the current environment. 

The proposed framework comprises offline parameter-error function modeling and online adaptive parameter selection. In the offline step, we run the odometry estimation algorithm $\mathcal{F}$ with different parameter samples $\bm{x}_j$ and environments $\mathcal{E}_i$ and evaluate the trajectory estimation errors $y_{ij}$. Subsequently, we build a surrogate function $\mathcal{S}$ that predicts the trajectory estimation error of the algorithm for a given parameter set and environment: $S({\bm x}_j, \mathcal{E}_i) \simeq y_{ij}$. In the online step, we identify the parameter set $\tilde{\bm{x}}_t$ that minimizes the predicted trajectory error for the current environment $\tilde{\mathcal{E}}_t$: $\tilde{\bm{x}}_t = \argmin_{\bm x} S (\bm{x}, \tilde{\mathcal{E}}_t)$.

We first demonstrate the role of the proposed framework in improving the accuracy of an odometry estimation algorithm across different environments through a simulation-based toy example. Further, an evaluation on the public dataset KITTI shows that the proposed framework can be used in real applications and improves the accuracy of odometry estimation algorithms with completely different architectures (Generalized ICP (GICP) odometry \cite{Segal2009}, LeGO-LOAM \cite{Shan2018}, and SuMa \cite{Behley2018}).

%%%%%%%%%%%%%%%%%%%%%%%%%%%%%%%%%%%%%%%%%%%%%%%%%%%%%%%%%%%%%%%%%%%%%%%%%%%%%%%%
\section{RELATED WORK}

SLAM and odometry estimation algorithms are inherently very complex and use many hyperparameters (e.g., point cloud resolution, feature matching threshold, and keyframe interval). It is known that, to achieve the best results, some popular SLAM frameworks such as Google Cartographer \cite{Hess2016} need to fine-tune many hyperparameters that affect each other\footnote{\url{https://google-cartographer-ros.readthedocs.io/en/latest/tuning.html}}. This tuning process requires a deep understanding of the inner working of the frameworks and considerable human effort of trial and error.

Despite the significance of hyperparameters, there are only a few studies on the automatic parameter tuning of SLAM-related algorithms. Nobili {\it et al.} automatically optimized the outlier removal parameters of ICP algorithms for a humanoid robot based on a metric of the overlap between point cloud pairs \cite{Nobili_2017}. Zheng used an exhaustive grid search to optimize the parameters of a visual SLAM algorithm (e.g., number of features, patch size, and edge threshold) \cite{Zheng_2020}. Similarly, Permeleau {\it et al.} investigated the changes in the ICP registration accuracy depending on the parameter values using exhaustive parameter-by-parameter evaluation \cite{Pomerleau_2011}. Although several techniques have been proposed for the fine-tuning of SLAM-related algorithms for specific use scenarios \cite{Pomerleau2015}, all the above works are dedicated to specific sensors, environments, and task configurations. 

In the context of machine learning, hyperparameter tuning has been recognized as an important step in maximizing the performance of learning models \cite{probst2019tunability}. When the number of parameters to be tuned is small, fine-tuning is often performed by exhaustive grid and random searches \cite{random_search}. They are easy to implement and work well for classic models that use only a few parameters (e.g., RBF SVM parameter selection \cite{6321759}). However, the required number of evaluations with these methods rapidly increases as the number of parameters increases. Thus, they are not suitable for tuning computationally expensive models with many parameters (e.g., deep neural networks). 

In recent years, the sequential model-based optimization (SMBO) approach has been used to efficiently optimize many parameters in computationally expensive models \cite{NIPS2011_4443, Shahriari2016}. This approach introduces a surrogate model that approximates the expensive model evaluation and samples the parameter set that maximizes an acquisition criterion (e.g., expected improvement) on the surrogate model. This facilitates the efficient exploration of the parameter space with fewer evaluations. Recent deep-learning-based methods, which require the optimization of many hyperparameters for an expensive training model, have benefited from SMBO-based automated hyperparameter tuning \cite{Liang2018}.

Although we can simply apply the SMBO approach to LiDAR odometry estimation such that the trajectory evaluation metric is minimized on training sequences \cite{koide_icra2021}, we observed that simple SMBO-based parameter optimization can considerably suffer from overfitting to the training sequences for two reasons. First, we typically have fewer training sequences in the odometry estimation problem compared to the usual machine learning problems. Second, the corruption of the trajectory estimation results in a catastrophic error in the trajectory evaluation metrics (e.g., absolute and relative trajectory errors \cite{Zhang2018}), which causes the SMBO to choose a very conservative parameter set. It is difficult for most odometry estimation algorithms to deal with different environments using a single fixed parameter set; hence, we argue that an adaptive mechanism to select an appropriate parameter set depending on the environment is necessary to further improve the trajectory estimation accuracy across different environments.

Although there are a few works that adaptively optimize the hyperparameters of specific algorithms depending on the environment (e.g., adaptive octree-resolution selection for visual SLAM \cite{Vespa2019}), to the best of our knowledge, no work has proposed a general and adaptive hyperparameter tuning framework without a limitation on the algorithms to be tuned.

%%%%%%%%%%%%%%%%%%%%%%%%%%%%%%%%%%%%%%%%%%%%%%%%%%%%%%%%%%%%%%%%%%%%%%%%%%%%%%%%
\section{METHODOLOGY}

Fig. \ref{fig:system} shows an overview of the proposed adaptive hyperparameter tuning framework for {\it black-box} odometry estimation algorithms. We first model a surrogate function $\mathcal{S}$ offline that predicts the trajectory estimation error for a given parameter set $\bm{x}$ and environment descriptor $\mathcal{E}$. To achieve this, we run the odometry estimation algorithm multiple times with different parameter samples $\bm{x}_j$ and environments $\mathcal{E}_i$ and evaluate the trajectory estimation errors $y_{ij}$. A k-nearest neighbor regressor-based surrogate function is then fitted to the evaluation results: $S(\bm{x}_j, \mathcal{E}_i) \simeq y_{ij}$. To efficiently explore the parameter space, we employ an SMBO-based parameter sampling technique. We then extract the environment descriptor $\tilde{\mathcal{E}}_t$ for the input point cloud $\tilde{\mathcal{P}}_t$ and identify the parameter set $\tilde{\bm{x}}_t$ that minimizes the predicted error with the surrogate function: $\tilde{\bm{x}}_t = \argmin_{\bm{x}} S(\tilde{\mathcal{E}}_t, \bm{x})$.

It should be noted that, while we assume uniform distributions for the parameters to be tuned for simplicity, the proposed framework can naturally be extended to accept other parameter distribution classes (e.g., discrete and categorical distributions).

%%%%%%%%%%%%%%%%%%%%%%%%%%%%%%%%%%%%%%%%%%%%%%%%%%%%%%
\subsection{Offline Parameter-Error Function Modeling}

For each training sequence $\mathcal{D}$, we run the odometry estimation algorithm $\mathcal{F}$ multiple times with different parameter samples $\bm{x}_j$ and obtain a set of estimated trajectories $\mathcal{T}_j = \mathcal{F}(\mathcal{D}, \bm{x}_j)$. The accuracy of each trajectory is evaluated using a metric $\mathcal{G}$, such as relative trajectory errors (RTEs) \cite{Zhang2018}. We use the notation $\mathcal{G}(\mathcal{F}(\mathcal{D}, \bm{x})) = \mathcal{G}(\bm{x})$ for brevity. To efficiently explore the parameter space, we employ the SMBO-based parameter sampling technique shown in Algorithm \ref{alg:smbo}. The SMBO fits a surrogate model $\mathcal{M}_j$ to the evaluation history $\mathcal{H}$ and samples a parameter set that maximizes the acquisition criterion $\mathcal{A}$ on the surrogate model. We use the expected improvement (EI) \cite{Zhan2020} as the acquisition criterion, with the expectation that $y$ negatively exceeds a threshold $y^*$:

\begin{align}
  \label{eq:ei}
  EI_{y^*} := \int^{\infty}_{-\infty} \max \left(y^* - y, 0 \right) p_{M_i}(y|{\bm x}) dy.
\end{align}
The parameter set $\bm{x}^*$ that maximizes EI can be determined by introducing two density functions $l({\bm x})$ and $h({\bm x})$, which are respectively fitted to the evaluation history where $y < y^*$ and $y >= y^*$, and identifying the parameter set that maximizes $g({\bm x}) / l({\bm x})$. $y^*$ is a threshold chosen such that the quantile $\gamma = p(y < y^*)$. We refer the reader to \cite{NIPS2011_4443} for a detailed derivation.

\begin{figure}[tb]
  \centering
  \includegraphics[width=\linewidth]{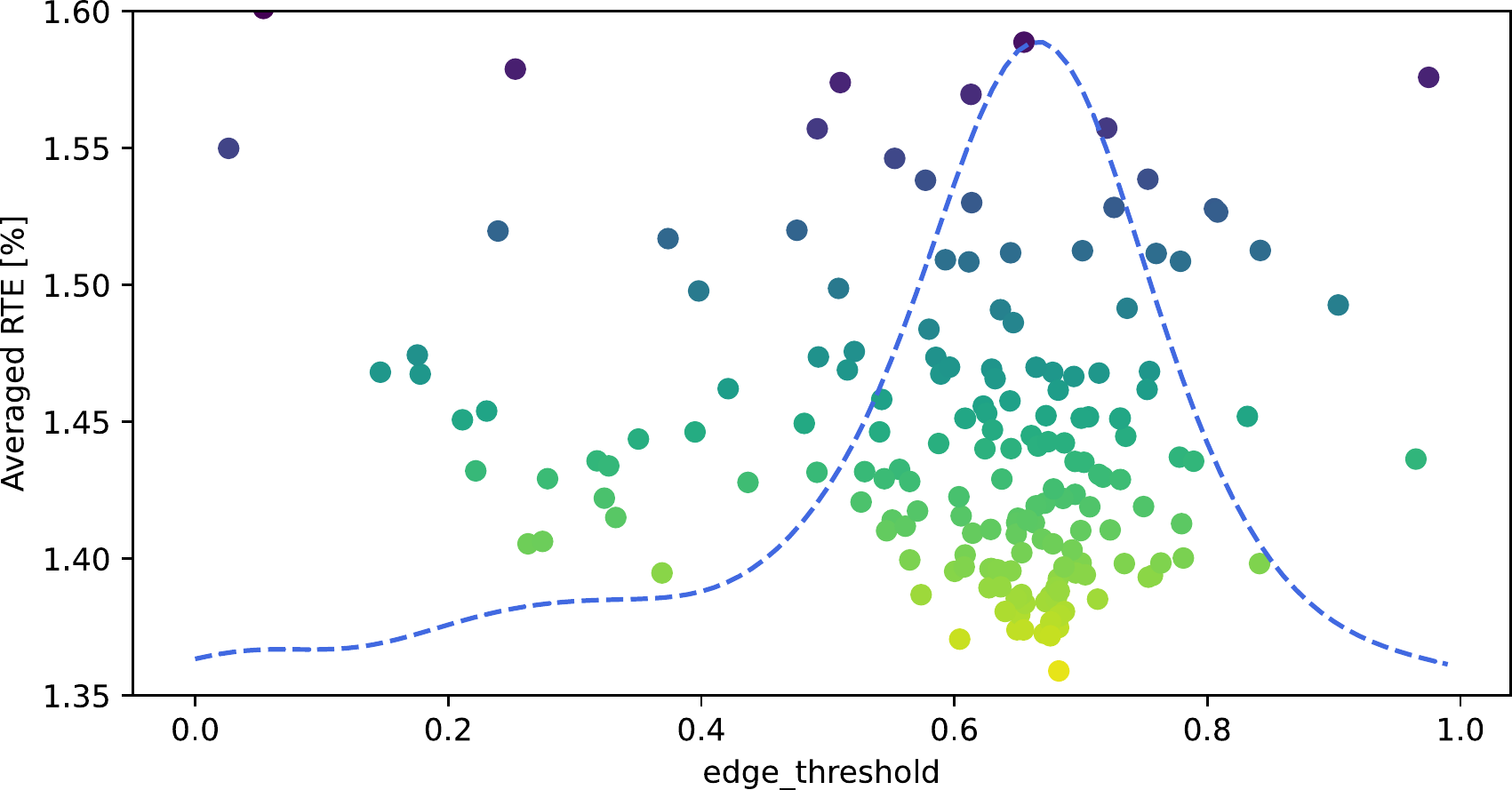}
  \caption{Example of parameters sampled with SMBO. The blue dashed line indicates the density of the sampled parameters. More parameters are sampled from the region where the error is expected to be small.}
  \label{fig:param_samples}
\end{figure}

For parameter sampling, the SMBO is used to minimize the average translational RTEs. We set the number of SMBO iterations to 256, and thus obtain 256 parameter sets and corresponding trajectories for each training sequence. Fig. \ref{fig:param_samples} shows example parameter values of LeGO-LOAM \cite{Shan2018} sampled with the SMBO. The blue line indicates the density of the sampled parameters given by the kernel density estimation. It can be observed that the SMBO efficiently explores the parameter space by taking more samples from the region where the error is expected to be small. 

For each frame $i$, we calculate the RTE $y_{ij}$ of the sub-trajectory starting from the frame with a parameter set $\bm{x}_j$. The parameter values $\bm{x}_j$ are then concatenated with the environment descriptor $\mathcal{E}_i$ of the starting frame $i$ to compose a feature vector ${\bm q}_{ij} = [\mathcal{E}_i, {\bm x}_j]$, which describes the combination of the environment and parameter values. We then model a surrogate function $\mathcal{S}$ that predicts the estimation error for a given parameter set and environment by fitting a k-nearest neighbor regressor (k = 5) such that $\mathcal{S}(\mathcal{E}_i, \bm{x}_j) \simeq y_{ij}$.

For simplicity, we use the descriptor of the first frame of a sub-trajectory to predict the trajectory estimation error. However, error prediction inherently requires information on all the frames in the time period of the sub-trajectory. Although this simplification can be justified by assuming that consecutive frames have similar environment structures in odometry estimation, extensions using a sequence of descriptors to capture drastic environmental changes and sensor motion can be considered in future work.

\begin{algorithm}[tb]
\caption{Sequential Model-Based Optimization \cite{NIPS2011_4443}}
\label{alg:smbo}
\begin{algorithmic}[1]
  \State $\mathcal{H} \leftarrow []$
  \For{$ j \in \left[1, \cdots, N \right]$}
    \State ${\bm x}^* \leftarrow \argmax_{{\bm x}} \mathcal{A}\left( \mathcal{M}_{j - 1} \left( {\bm x} \right) \right) $
    \State $y^* \leftarrow \mathcal{G} \left( {\bm x}^* \right)$
    \State $\mathcal{H} \leftarrow \mathcal{H} \cup \left( {\bm x}^*, y^* \right)$
    \State Fit $\mathcal{M}_j$ to $\mathcal{H}$
  \EndFor
  \Return $\mathcal{H}$
\end{algorithmic}
\end{algorithm}

%%%%%%%%%%%%%%%%%%%%%%%%%%%%%%%%%%%%%%%%%%%%%%%%%%%%%%
\subsection{Online Parameter Selection}

For the online parameter selection, we first extract the environment descriptor $\tilde{\mathcal{E}}_t$ for the input point cloud $\tilde{\mathcal{P}}_t$ and find the parameter set $\tilde{\bm{x}}_t$ that minimizes the predicted error with the surrogate function:

\begin{align}
\tilde{{\bm x}}_t = \argmin_{\bm x} \mathcal{S}(\tilde{\mathcal{E}}_t, {\bm x}).
\end{align}

Because the surrogate function $\mathcal{S}$ is a k-nearest neighbor regressor that is nonlinear and non-convex, its minimization is not trivial. Thus, we again use the SMBO parameter search to determine the best parameter set $\tilde{\bm x}_t$ for the current environment. We perform parameter selection every second and update the parameter set.

%%%%%%%%%%%%%%%%%%%%%%%%%%%%%%%%%%%%%%%%%%%%%%%%%%%%%%
\subsection{Environment Descriptor}

As a descriptor to represent the environment structure, we use simple NDT (Normal Distributions Transform) voxel classification-based features \cite{Magnusson2009}. Following \cite{Magnusson2009}, we calculate the normal distribution voxels for the input point cloud. Based on the eigenvalues of the covariance matrices ($\lambda_1 > \lambda_2 > \lambda_3$), we classify the voxels into linear ($\lambda_2 / \lambda_1 < t_e$), planar ($\lambda_3 / \lambda_2 < t_e$), and spherical (other cases) distributions; $t_e$ is a constant in $(0, 1)$. Each distribution class has three subclasses based on the line tilt angle, plane tilt angle, and eigenvalue ratio ($\lambda_1 / \lambda_2$), and they are invariant to yaw rotation. We count the number of each subclass for several distance ranges and create a histogram (3.0 m $\times$ 10 bins $\times$ 9 sub-classes). We then apply PCA dimensionality reduction ($D = 10$) to obtain the feature vector $\mathcal{E}$ representing the environment structure.

It is worth mentioning that the proposed framework is agnostic to the environment descriptor, and other features (e.g., hand-crafted \cite{Rusu2009} as well as learned features \cite{Gojcic2019}) can be used with slight modifications.

%%%%%%%%%%%%%%%%%%%%%%%%%%%%%%%%%%%%%%%%%%%%%%%%%%%%%%%%%%%%%%%%%%%%%%%%%%%%%%%%
\section{Experiments}

%%%%%%%%%%%%%%%%%%%%%%%%%%%%%%%%%%%%%%%%%%%%%%%%%%%%%%
\subsection{Simulation-based Toy Example}

\begin{figure}[tb]
  \centering
  \includegraphics[width=\linewidth]{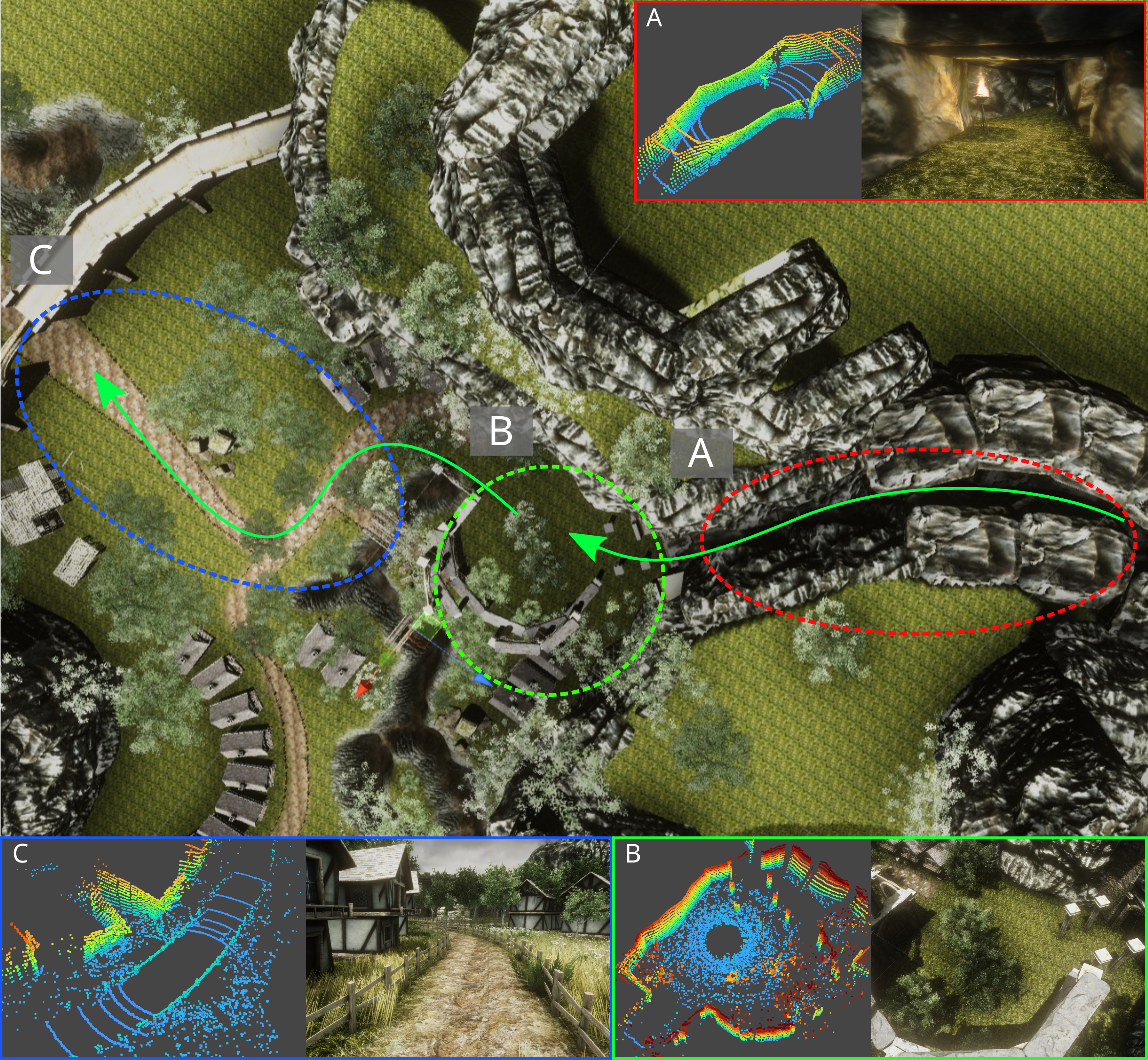}
  \caption{Simulation environment for the toy example. The robot moves across cave (A) and outdoor (C) environments through an open space (B). These environments have very different structures and thus require different parameter settings.}
  \label{fig:simenv}
\end{figure}

\begin{figure}[tb]
  \centering
  \includegraphics[width=0.95\linewidth]{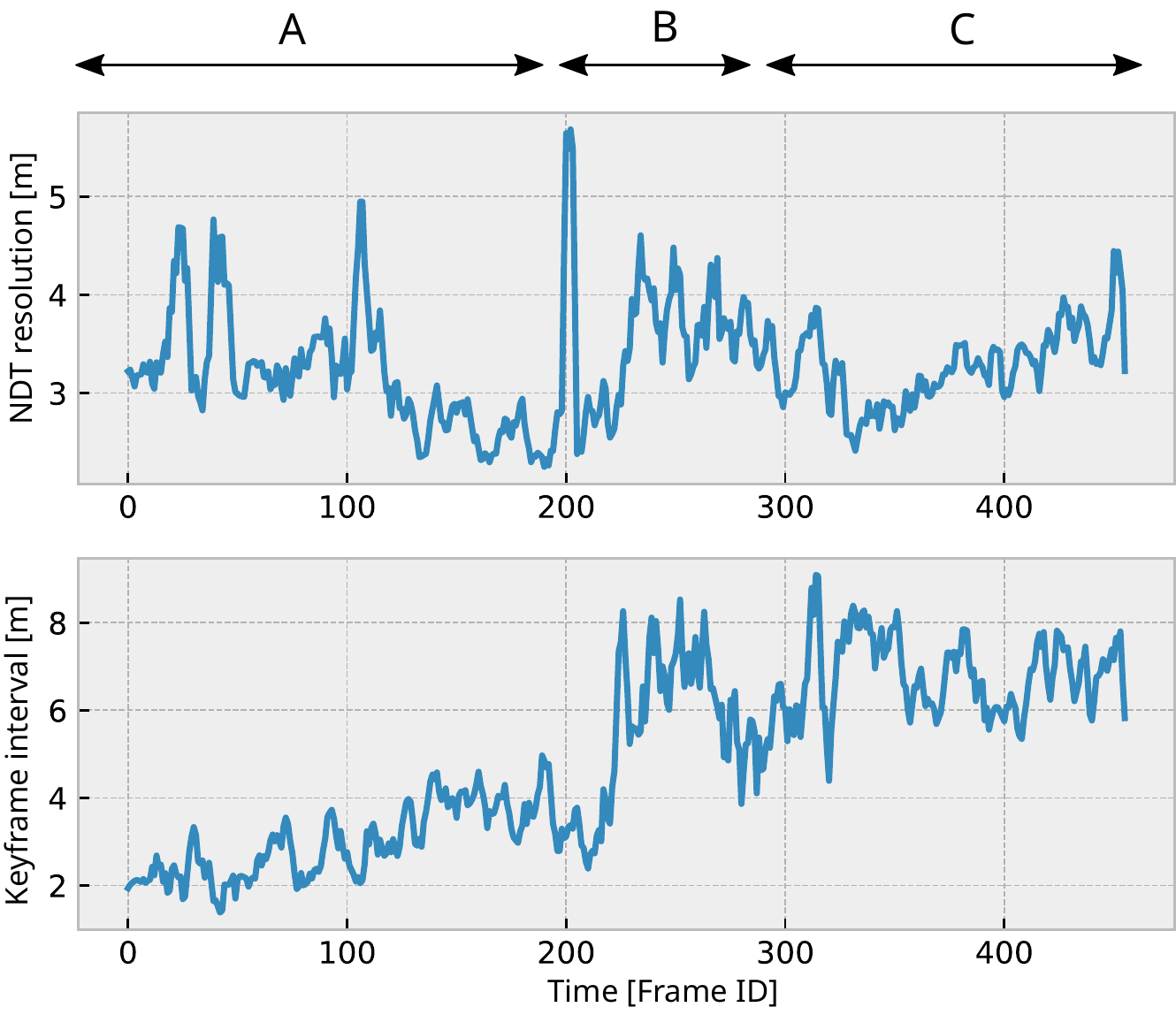}
  \caption{Adaptively selected parameters through different environments in the toy example.}
  \label{fig:adaptive}
\end{figure}

To demonstrate the process where the proposed framework adaptively selects parameters and improves the accuracy of the odometry estimation across different environments, we conducted experiments in a simulated environment, as shown in Fig. \ref{fig:simenv}. The environment consists of a cave (A), open space (B), and outdoor street environments (C). They have very different environmental structures and thus require different parameter settings. We generate LiDAR point clouds at 10 Hz with the sensor configuration of Velodyne HDL32e.

We used a simple keyframe-based NDT scan matching odometry algorithm with only two parameters: NDT voxel resolution and keyframe update interval. This uses the NDT algorithm with the specified voxel resolution value, and updates the keyframe (registration target point cloud) every time the travelled distance from the last keyframe exceeds the keyframe interval threshold parameter. These parameters provide a trade-off between accuracy and stability (large voxel size: better convergence and lower accuracy, large keyframe interval: low drift and worse stability) and need to be tuned depending on the environment.

For offline parameter-error function modeling, we recorded four LiDAR data sequences and ground truth trajectories for each of the cave and outdoor environments. For each sequence, we sampled 256 parameter sets and calculated the translational RTEs of 25 m sub-trajectories to evaluate the estimated trajectories. For validation, we recorded a sequence across the cave, open space, and outdoor environments.

We ran the NDT scan matching odometry algorithm on the validation sequence with three parameter settings: 1) manually tuned baseline parameter set, 2) fixed parameter set tuned with SMBO for the training sequences, and 3) adaptively selected parameter sets. Fig. \ref{fig:adaptive} shows a plot of the adaptively selected NDT resolution and keyframe interval parameters. It can be observed that small keyframe interval values were selected to prevent estimation corruption in the cave environment (A), whereas large interval values were selected after the sensor moved out from the cave to improve the accuracy (B)(C). It can also be observed that large NDT resolution values tend to be selected in (B) and (C) to better capture the structure of the open environments. It is worth emphasizing that the proposed framework automatically determines how the parameters should be tuned for different environments using the data-driven approach; consequently, it does not require detailed information on the inner working of the algorithm to be tuned.

\begin{table}[tb]
  \centering
  \caption{Translational RTEs [\%] on the toy example sequence with different parameter sets}
  \label{tab:error_toy}
  \begin{tabular}{c|c|c}
Method    & Parameter setting & RTE                 \\ \hline \hline
\multirow{3}{*}{NDT odometry}
          & Baseline          & 0.767                \\
          & Tuned (Fixed)     & 0.759 (-0.008)       \\
          & Tuned (Adaptive)  & {\bf 0.530 (-0.237)} \\
  \end{tabular}

Values in parentheses are with respect to the baseline
\end{table}

Table \ref{tab:error_toy} shows the averaged translational RTEs for the different parameter settings. Although the fixed tuned parameter set exhibited slightly better accuracy (0.759 \%) compared to the manually tuned baseline (0.767 \%), the improvement was small. This is because it needed to select a conservative parameter set for the cave environment, which did not result in accuracy improvement in the outdoor environment. In contrast, the adaptive parameter selection significantly improved the estimation accuracy (0.530 \%). This is because it adaptively changed the parameter set from conservative to aggressive after leaving the cave environment.

This simple toy example illustrates the necessity of an adaptive parameter selection mechanism to enhance the trajectory estimation accuracy across different environments.

%%%%%%%%%%%%%%%%%%%%%%%%%%%%%%%%%%%%%%%%%%%%%%%%%%%%%%
\subsection{Evaluation on KITTI}

\begin{figure}[tb]
  \centering
  \includegraphics[width=\linewidth]{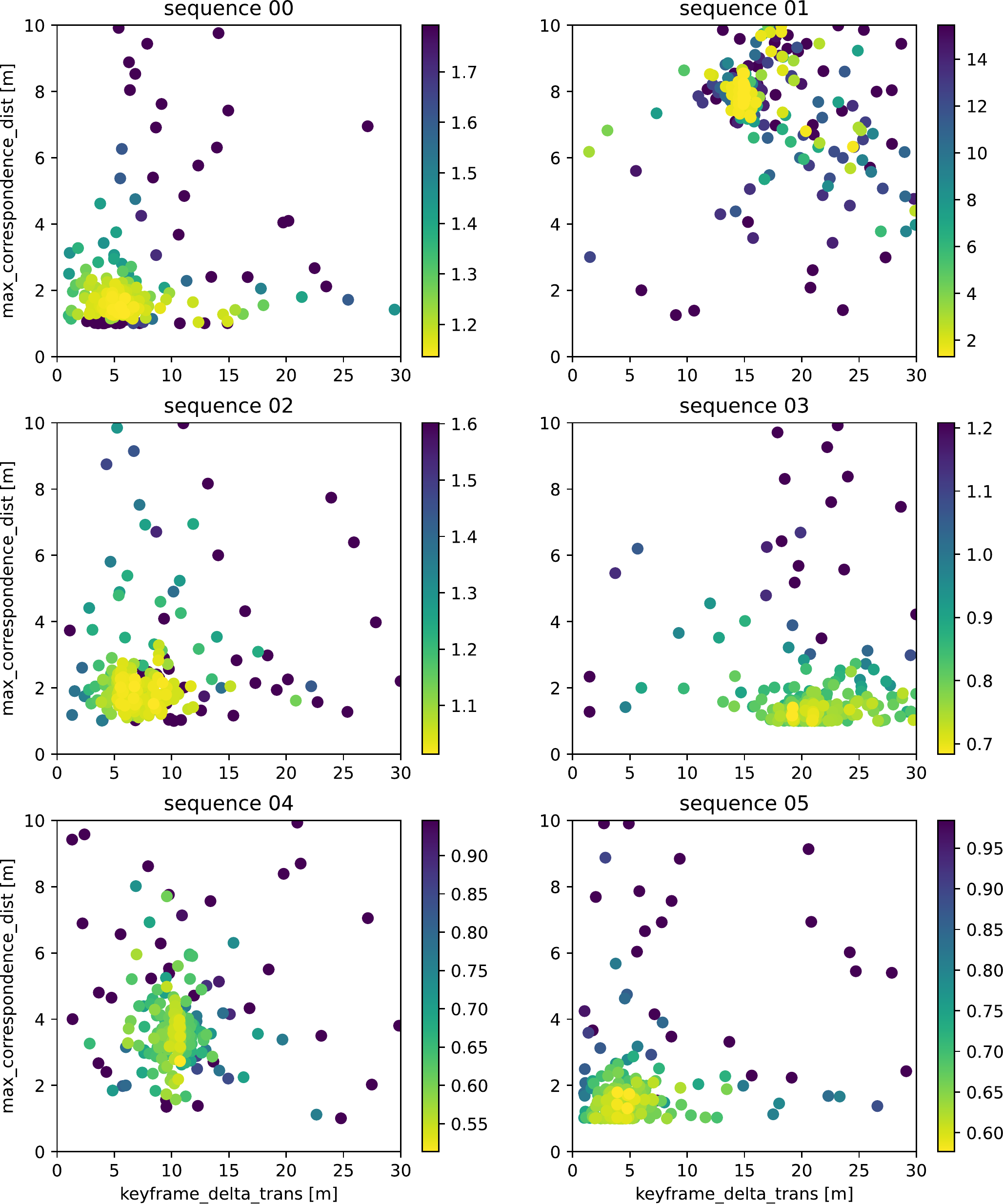}
  \caption{Parameters of GICP scan matching odometry sampled with SMBO for KITTI sequence 00 to 06. The color of the points indicates translational RTE [\%].}
  \label{fig:gicp_kitti}
\end{figure}

Next, we conducted an evaluation on a public LiDAR odometry dataset, KITTI, to demonstrate that the proposed approach improves the accuracy of several odometry estimation algorithms in practical situations. The dataset contained 11 sequences recorded using Velodyne HDL-64E at a rate of 10 Hz. We used sequences 00-05 for the parameter optimization, and sequences 06-10 for validation.

We evaluated three odometry estimation algorithms with different architectures: keyframe-based GICP odometry \cite{Segal2009}, LeGO-LOAM \cite{Shan2018}, and SuMa \cite{Behley2018} \footnote{The list of tuned parameters of each algorithm is available at\url{https://github.com/SMRT-AIST/automatic_tuning/blob/devel/adaptive_parameters.md}}. We added slight modifications to LeGO-LOAM and SuMa to dynamically update the hyper-parameters via ROS param mechanism. As the baseline, we manually tuned the parameter sets of the GICP odometry and LeGO-LOAM for sequence 00. For SuMa, we used the parameter set optimized for all the sequences in \cite{Behley2018}. We compared the trajectory errors of these algorithms with 1) the manually tuned baseline parameter set, 2) a fixed parameter set optimized with SMBO for the entire training sequences, and 3) adaptively selected parameter sets (the parameter-error function is modeled on the training set offline, and the parameters are adaptively selected on each sequence online). For the parameter-error function modeling of the adaptive parameter selection, we used translational RTEs of 100 m sub-trajectories as the evaluation metric.

The offline parameter-error function modeling terminated in about six hours for each odometry estimation method. The online feature extraction and parameter optimization, which were executed every second, took 40 msec and 700 msec, respectively. 

\begin{figure}[tb]
  \centering
  \begin{minipage}[b]{0.42\linewidth}
  \centering
  \includegraphics[height=5.4cm]{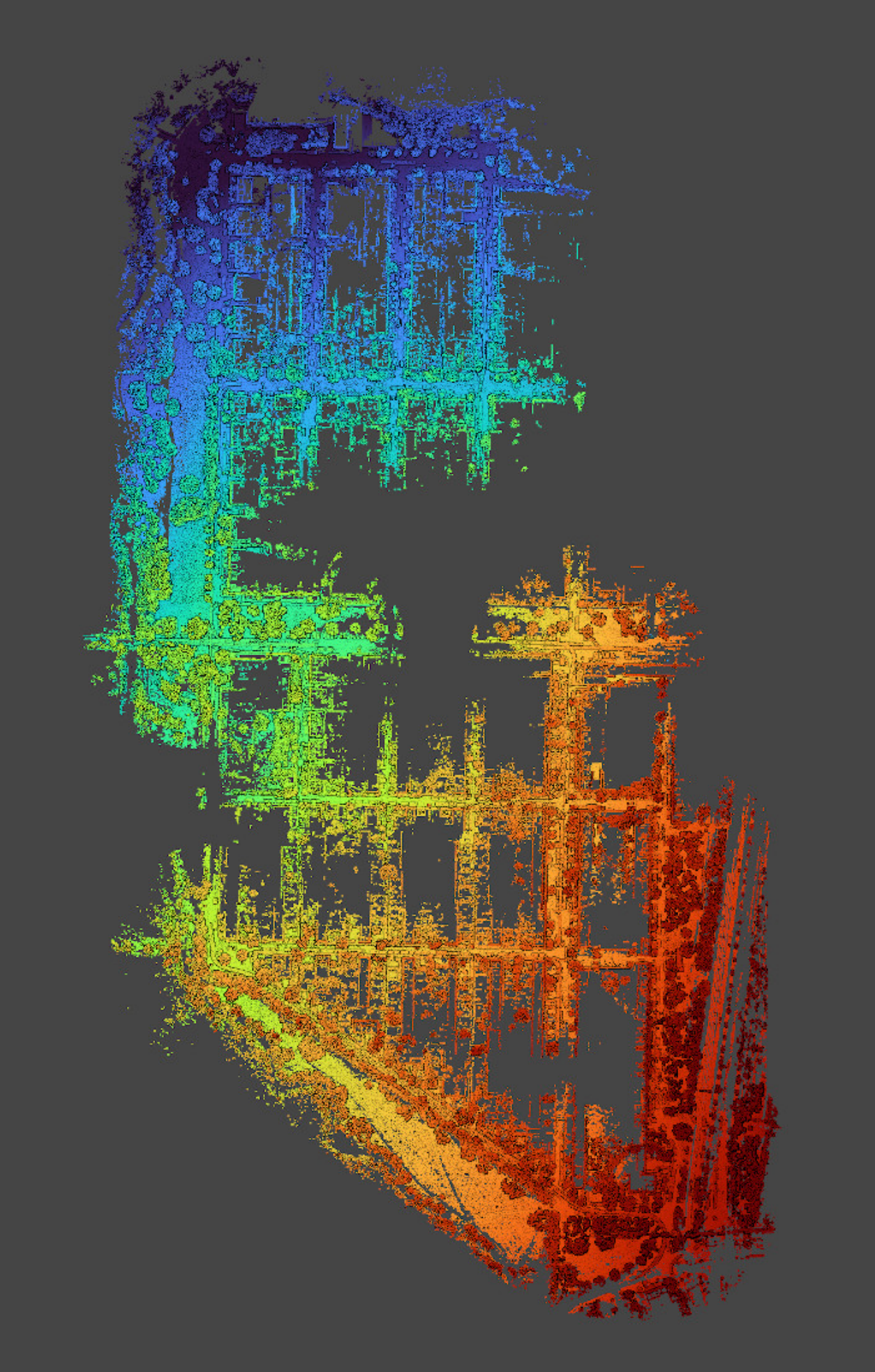}
  \subcaption{Environment (sequence 08)}
  \end{minipage}
  \begin{minipage}[b]{0.46\linewidth}
  \centering
  \includegraphics[height=5.6cm]{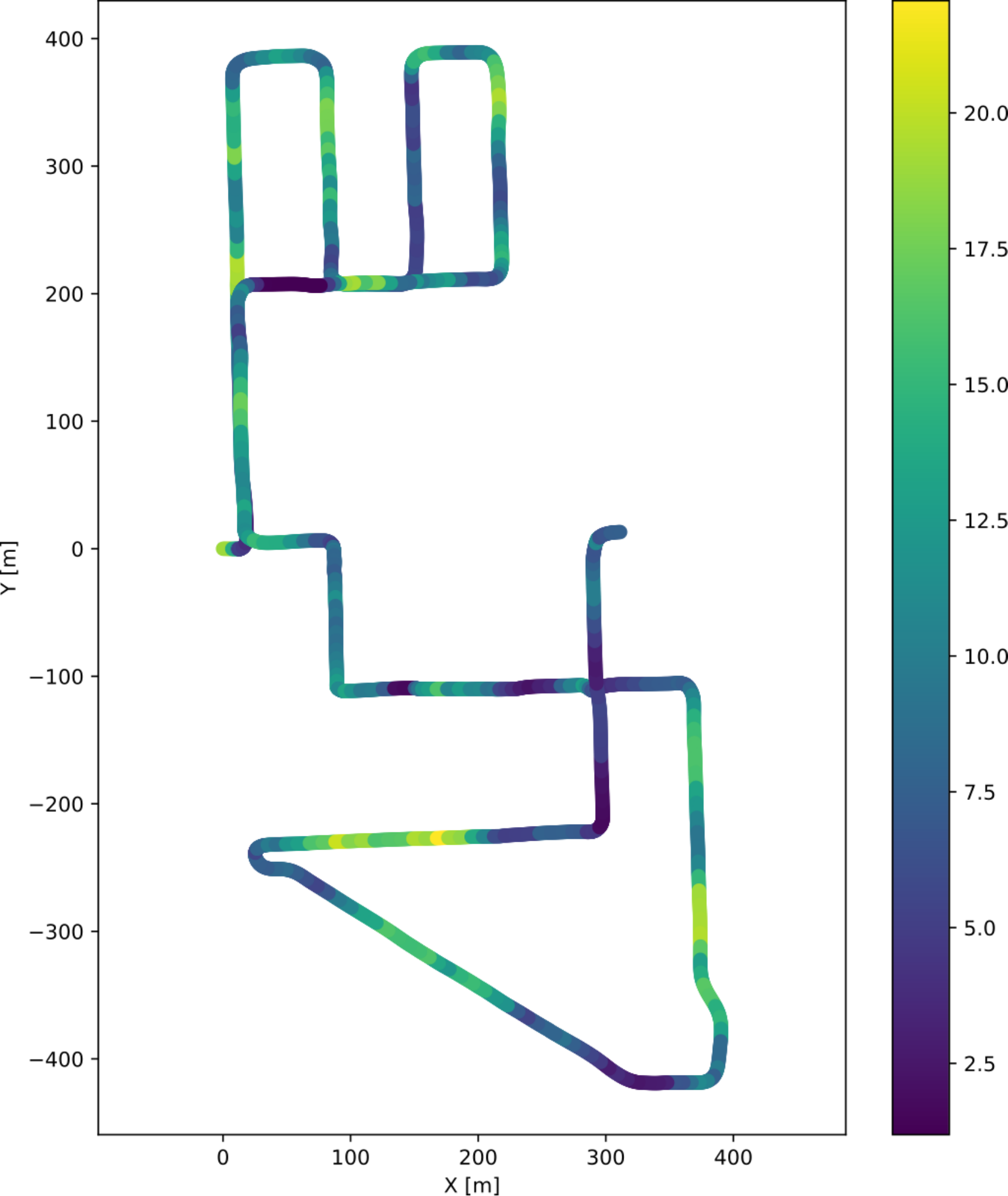}
  \subcaption{Keyframe interval}
  \end{minipage}
  \caption{Example of adaptively optimized hyperparameters.}
  \label{fig:seq08}
\end{figure}

\begin{table*}[tb]
  \centering
  \caption{Average translational RTEs [\%] for KITTI}
  \label{tab:error_kitti}
  \begin{tabular}{c||c|cccccc|c||ccccc|c}
\multirow{3}{*}{Method}
          &                    & \multicolumn{7}{c||}{Training set}                                                                         & \multicolumn{6}{c}{Test set}                                               \\
          & Seq. ID            & 00          & 01           & 02          & 03          & 04          & 05          & \multirow{2}{*}{Avg.} & 06          & 07          & 08          & 09          & 10          & \multirow{2}{*}{Avg.}  \\ 
          & \# Frames          & 4541        & 1101         & 4661        & 801         & 271         & 2761        &                       & 1101        & 1101        & 4071        & 1591        & 1201        &       \\ \hline \hline
\multirow{3}{*}{GICP}
          & Baseline           & 1.30         & 2.42         & 1.72         & 1.08         & 1.89         & 0.90         & 1.46               & 0.78         & 1.30         & 1.43         & 1.17         & 1.72         & 1.33               \\
          & Fixed              & {\bf 1.21}   & 2.45         & 1.66         & 1.08         & 0.99         & {\bf 0.61}   & 1.34 (-0.12)       & {\bf 0.55}   & 1.35         & 2.00         & {\bf 1.06}   & {\bf 1.12}   & 1.54 (+0.21)       \\
          & Adaptive           & {\bf 1.21}   & {\bf 2.16}   & {\bf 1.48}   & {\bf 0.97}   & {\bf 0.78}   & 0.64         & {\bf 1.26 (-0.20)} & 0.57         & {\bf 0.92}   & {\bf 1.18}   & 1.22         & 1.50         & {\bf 1.13 (-0.20)} \\
\hline
\multirow{3}{*}{LeGO-LOAM}
          & Baseline           & 1.73         & 13.33        & 1.76         & 1.61         & 1.35         & 1.07         & 2.44               & 1.05         & {\bf 0.97}   & 1.69         & 1.33         & 1.77         & 1.52               \\
          & Fixed              & 1.87         & 3.07         & 1.83         & 1.63         & 1.33         & 1.02         & 1.77 (-0.67)       & 1.09         & 1.02         & 1.76         & 1.29         & 1.83         & 1.56 (+0.04)       \\
          & Adaptive           & {\bf 1.47}   & {\bf 2.78}   & {\bf 1.53}   & {\bf 1.22}   & {\bf 1.21}   & {\bf 0.79}   & {\bf 1.45 (-0.99)} & {\bf 0.90}   & 1.02         & {\bf 1.58}   & {\bf 1.28}   & {\bf 1.73}   & {\bf 1.43 (-0.09)} \\
\hline
\multirow{3}{*}{SuMa}
          & Baseline           & 0.89         & 6.76         & 1.37         & 1.37         & {\bf 0.48}   & 1.06         & 1.52               & 0.62         & 0.70         & {\bf 1.36}   & 1.06         & 1.93         & 1.23               \\
          & Fixed              & {\bf 0.84}   & {\bf 5.59}   & 1.27         & {\bf 1.14}   & 0.56         & {\bf 0.87}   & {\bf 1.35 (-0.17)} & {\bf 0.60}   & 0.66         & 1.43         & {\bf 0.86}   & 1.73         & 1.21 (-0.02)       \\
          & Adaptive           & 0.90         & 5.69         & {\bf 1.24}   & 1.19         & 0.58         & 1.00         & 1.39 (-0.13)       & 0.61         & {\bf 0.65}   & 1.40         & 0.94         & {\bf 1.65}   & {\bf 1.20 (-0.03)} \\
  \end{tabular}

Values in parentheses are with respect to the baseline
\end{table*}

Fig. \ref{fig:gicp_kitti} shows the parameters of the GICP odometry (maximum correspondence distance and keyframe interval) for each training sequence (sequence 00-05) sampled during the offline parameter-error function modeling. The color of the points indicates the translational RTE. It can be observed that different sequences require very different parameters. In particular, sequence 01, which is the most difficult sequence in KITTI owing to the feature-less environment structure, requires a very large maximum correspondence distance parameter to prevent estimation corruption. It can also be observed that the required keyframe interval is significantly changed for each sequence. This result clearly demonstrates the necessity of the adaptive parameter tuning mechanism. If a conservative parameter set is chosen for the difficult sequence (sequence 01), the estimation results of other sequences would deteriorate. Conversely, if a parameter set for another sequence (e.g., sequence 00) is chosen, the estimation would be corrupted on the difficult sequence. There is no parameter set that works well for all the sequences, and thus the parameter set must be adaptively selected depending on the environment.

Table \ref{tab:error_kitti} shows the translational RTEs averaged over 100 to 800 m trajectories for the odometry estimation algorithms with different parameter settings. We used the KITTI official evaluation code to calculate the RTEs.

Although the fixed parameter set optimized with SMBO improved the RTEs of the GICP odometry for the training sequences, the RTEs on the test set deteriorated compared to the baseline. This is because the fixed parameter selection strategy required to choose a conservative parameter set for the difficult sequences to improve the RTEs on the training set. The proposed approach significantly improved the RTEs on both the training and test sets by adaptively selecting the best parameter sets depending on the environment. Fig. \ref{fig:seq08} shows the adaptively selected keyframe interval values of the GICP odometry for sequence 08. Small interval values were selected on corners to prevent estimation corruption, whereas large values were selected on several long straight paths to reduce the odometry drift and achieve better accuracy.

The results of LeGO-LOAM exhibited a similar trend to that of the GICP odometry. Although the fixed parameter set significantly improved the translational RTEs on the training set, the RTEs on the test set deteriorated. The adaptive parameter selection successfully improved the RTEs for both the training and test sets.

Although both the fixed and adaptive parameter sets significantly improved the RTEs of SuMa on the training set, the improvement on the test set was very small. This may be because 1) the baseline parameter set was highly optimized for the entire dataset, and 2) the modified parameters affected the consistency of the surfel map of SuMa, resulting in better accuracy in some test sequences (sequences 06, 07, 09, and 10) but worse accuracy in a long sequence (sequence 08). These results indicate the shortcomings of the current form of the proposed framework; it uses a simple environment descriptor and regression model that may not capture the fine behavior of the odometry estimation algorithm. We infer that this can be further improved by considering as input environment descriptor sequences to better represent the changing environmental structures and introducing a small amount of prior information on the relationship between the parameters and odometry estimation algorithm behavior.

%%%%%%%%%%%%%%%%%%%%%%%%%%%%%%%%%%%%%%%%%%%%%%%%%%%%%%
\section{Conclusions}

This study proposed an automatic and adaptive hyperparameter tuning framework for {\it black-box} LiDAR odometry estimation algorithms. The proposed method first runs the odometry estimation algorithm to be tuned with different parameters and environments offline, and creates a surrogate function to predict the estimation error for novel parameters and environments. For online parameter selection, it chooses the parameter set that is expected to yield better accuracy in the given environment with the surrogate function. We employed an SMBO technique for efficient offline parameter space exploration and online surrogate function minimization. The evaluation results showed that the proposed approach improves the accuracy of the odometry estimation algorithms without detailed information on their inner working.

While it has been shown that the proposed framework enables to adaptively optimize hyper-parameters across different environments, it is still challenging to select good parameters in novel environments. We plan to further evaluate and enhance the generality of the adaptively tuning framework with, e.g., data augmentation and regularization techniques.

%%%%%%%%%%%%%%%%%%%%%%%%%%%%%%%%%%%%%%%%%%%%%%%%%%%%%%%%%%%%%%%%%%%%%%%%%%%%%%%%
\section*{ACKNOWLEDGMENT}

This work was supported in part by a project commissioned by the New Energy and Industrial Technology Development Organization (NEDO).

\balance

%%%%%%%%%%%%%%%%%%%%%%%%%%%%%%%%%%%%%%%%%%%%%%%%%%%%%%%%%%%%%%%%%%%%%%%%%%%%%%%%
\bibliography{iros2021}

\end{document}